\documentclass[a4paper]{article}
\usepackage{INTERSPEECH2021}
\usepackage{pifont}
\usepackage{multirow}
\usepackage{tabularx}
\usepackage{amsmath}
\usepackage{graphicx}
\usepackage[T1]{fontenc}
\usepackage{scalerel,xparse}

\newcommand{\ie}{\emph{i.e.}}
\newcommand{\etal}{\emph{et~al.}}
\newcommand{\bertshare}{{\sc{bertShare}}}
\newcommand{\bertbert}{{\sc{bert2bert}}}
\newcommand{\bertgpt}{{\sc{bert2gpt2}}}
\NewDocumentCommand\emojicat{}{\includegraphics[scale=0.03]{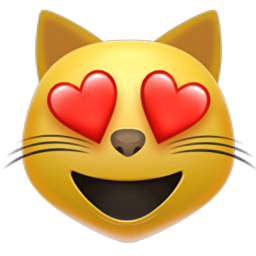}}

\begin{document}
\title{Enhancing Semantic Understanding with Self-supervised Methods\\for Abstractive Dialogue Summarization}

\name{Hyunjae Lee,\,Jaewoong Yun,\,Hyunjin Choi,\,Seongho Joe,\,Youngjune L. Gwon}
\address{AI Research Center, Samsung SDS, South Korea}
\email{\{h8.lee, jw0531.yun, hjjin.choi, drizzle.cho, gyj.gwon\}@samsung.com}

\maketitle

\begin{abstract}
Contextualized word embeddings can lead to state-of-the-art performances in natural language understanding. Recently, a pre-trained deep contextualized text encoder such as BERT has shown its potential in improving natural language tasks including abstractive summarization. Existing approaches in dialogue summarization focus on incorporating a large language model into summarization task trained on large-scale corpora consisting of news articles rather than dialogues of multiple speakers. In this paper, we introduce self-supervised methods to compensate shortcomings to train a dialogue summarization model. Our principle is to detect incoherent information flows using pretext dialogue text to enhance BERT's ability to contextualize the dialogue text representations. We build and fine-tune an abstractive dialogue summarization model on a shared encoder-decoder architecture using the enhanced BERT. We empirically evaluate our abstractive dialogue summarizer with the SAMSum corpus, a recently introduced dataset with abstractive dialogue summaries. All of our methods have contributed improvements to abstractive summary measured in ROUGE scores. Through an extensive ablation study, we also present a sensitivity analysis to critical model hyperparameters, probabilities of switching utterances and masking interlocutors. 
\end{abstract}


\section{Introduction}
In natural language processing, abstractive summarization generates a concise summary for lengthy source text using words that do not necessarily appear in the source. Such creative aspect (in comparison with extractive summarization) makes abstractive summarization one of the most challenging tasks in computational linguistics. In speech, dialogue summarization enables a useful capability to capture salient information scattered in a dialogue containing the utterances by multiple interlocutors and rewrite them into simplified, easy-to-grasp text. With the rapid growth of online communications, providing dialogue summaries is becoming one of the most important features in a speech system. The recent outbreak of the coronavirus pandemic (COVID-19) and other on-going global incidents demand a crisp summary of long speech conversations more useful and appealing than ever.

Self-supervised learning has been used widely to complement or replace entirely human-annotated datasets in training deep networks of language, speech, and visual models. There are numerous approaches for neural dialogue summarization, yet not many have aimed to improve semantic and structural understanding of dialogue with a specifically-designed self-supervised learning method.

In this paper, we propose self-supervised methods for training a neural abstractive dialogue summarization model. We have designed pretext tasks that require the model under training predict whether there are incorrect ordering and irrelevant information in utterances or not. There are also tasks to predict switched and masked interlocutor names. When pre-training with our self-supervised methods, the model seems to learn a better understanding of the semantic relevance between interlocutor and utterance. This can be crucial to capture the essence of a whole dialogue in much shortened text while preserving the most salient information.  

\begin{figure}[t!]
  \centering
  \includegraphics[width=\linewidth]{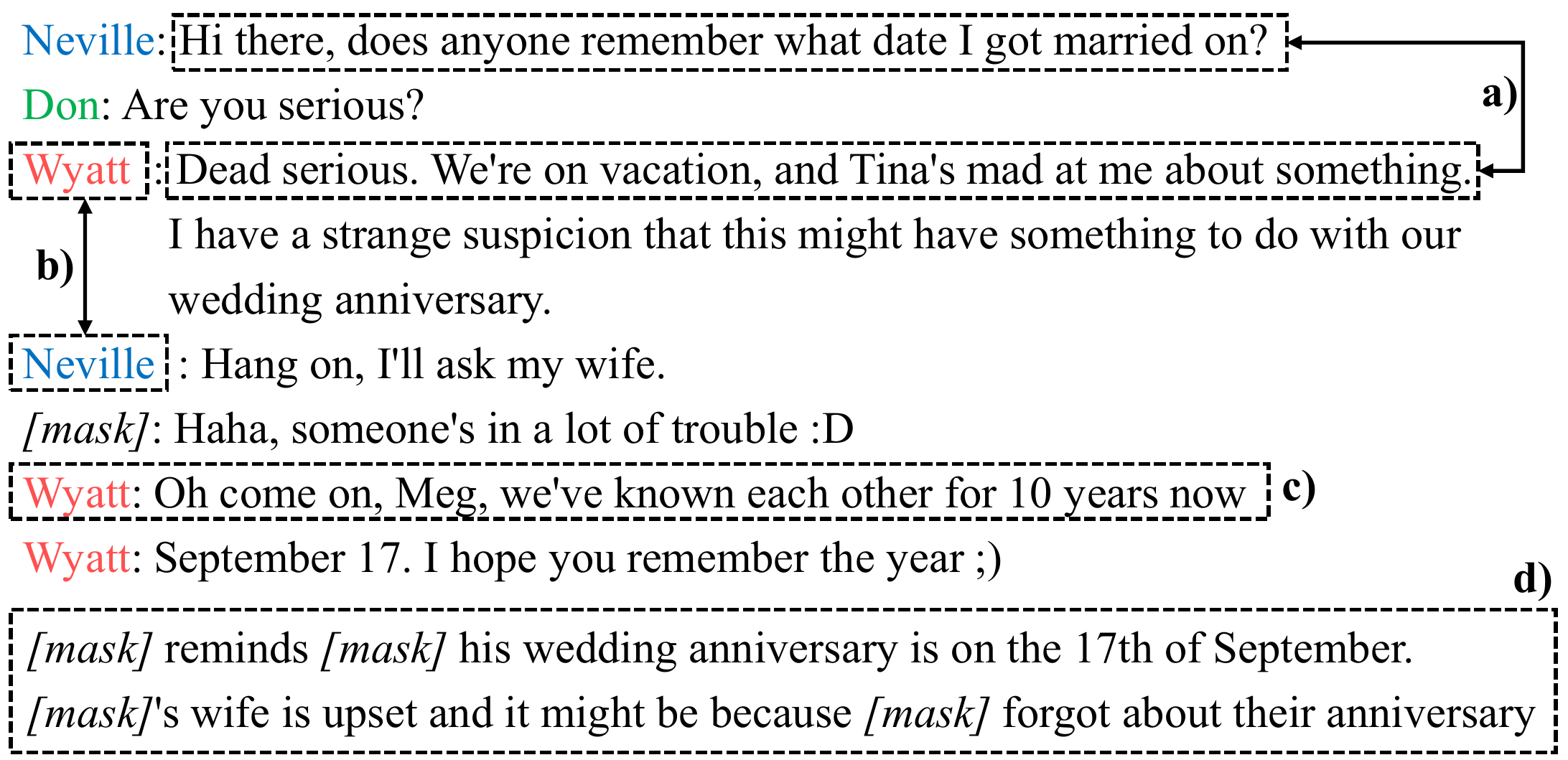}
  \caption{An overview of proposed self-supervised methods. They are \textbf{a) switching utterance}, \textbf{b) switching interlocutor}, \textbf{c) inserting utterance}, \textbf{d) masking interlocutor} methods. Each method is carried out separately. (Note that some methods can be combined to run beneficially.)}
  \label{fig:methods}
\end{figure}
 
In Figure~\ref{fig:methods}, we describe an illustrative example about how our self-supervised methods take place during the pre-training of a neural language model upon which abstractive summarization task can be fine-tuned. There are four self-supervised methods, namely switching utterance, switching interlocutor, inserting utterance, and masking interlocutor. Notice that three out of the four self-supervised methods are set up as a simple binary classification problem (\ie, corrupted or not). In masking interlocutor method, mapping to a correct interlocutor's name would be required instead. We remark that our proposed method is fully compatible with a publicly available pre-trained language model like BERT~\cite{devlin-etal-2019-bert}. This makes our approach more appealing and retrofitting in the popular paradigm comprising the pre-training and fine-tuning stages for building contemporary NLP applications.

\begin{figure}[t]
\begin{center}
\includegraphics[width=1.0\linewidth]{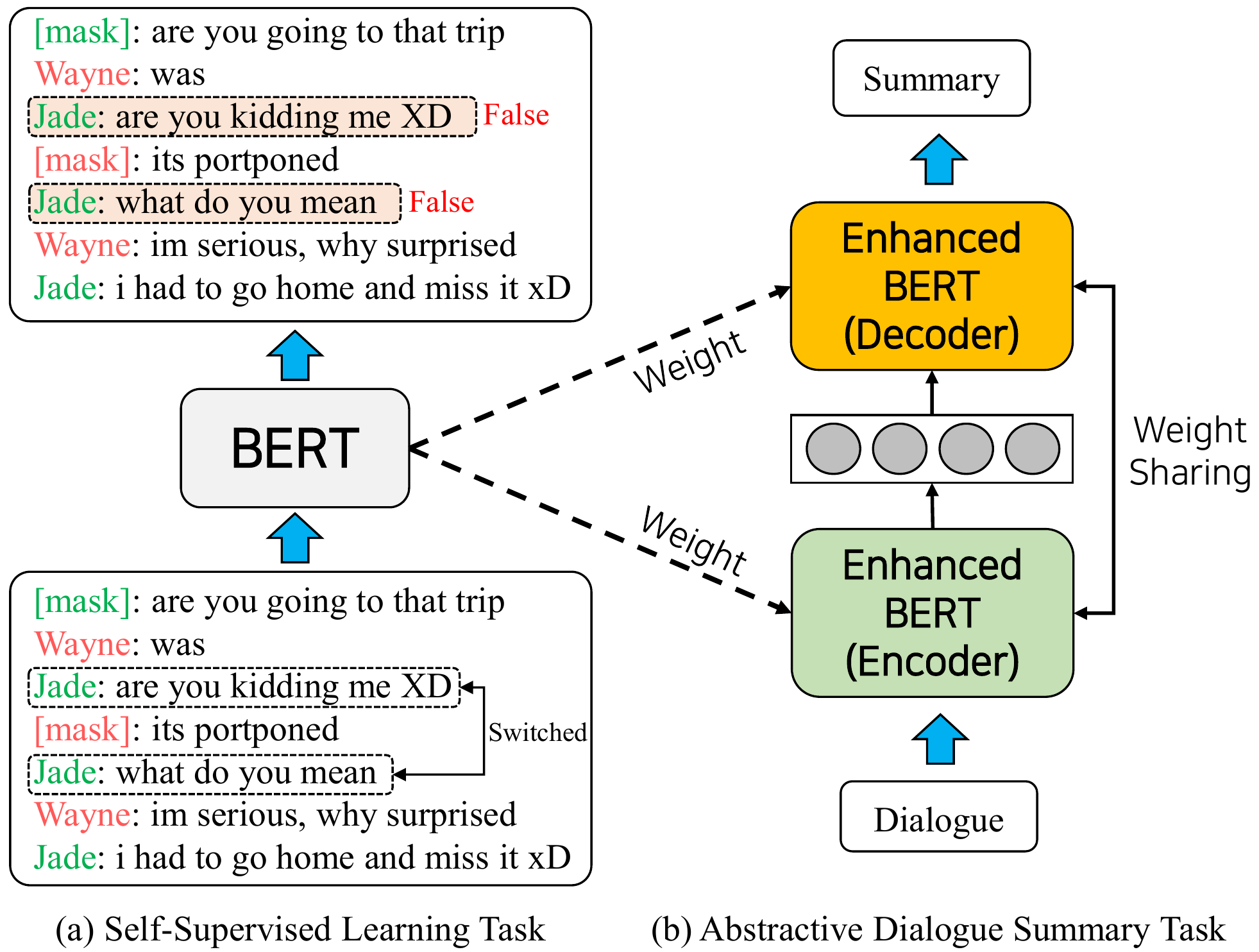}
\end{center}
\caption{Illustration of our approach on abstractive summarization task. First, we enhance dialogue context understanding of BERT via (a) proposed self-supervised methods. Then, we initialize the traditional encoder-decoder model with enhanced BERT and fine-tune on abstractive summarization task.}
\label{fig:archi}
\end{figure}

To construct a neural abstractive dialogue summarizer, we enhance a pre-trained BERT with our self-supervised methods and use it as both encoder and decoder in sequence-to-sequence model~\cite{Bahdanau2015NeuralMT, rothe-etal-2020-leveraging} while sharing the weights between the encoder and the decoder (see Figure~\ref{fig:archi}). We fine-tune and evaluate empirically our BERT-based summarizer using the SAMSum corpus~\cite{gliwa-etal-2019-samsum}. Our self-supervised methods indicate a significantly improved performance compared to the baseline (\bertshare~), which is using the pre-trained BERT as is (without applying the proposed self-supervised methods).

\section{Related Work}
\subsection{Self-supervised pre-training for text summarization}
In recent years, self-supervised learning has pushed the performance of a wide range of natural language processing (NLP) tasks to new state-of-the-art and becoming a dominant paradigm in NLP. Numerous pre-training methods based on self-supervised learning for text summarization have been introduced. A common approach lets a model predict the original input tokens from randomly masked tokens (or sentences) in a document to resemble a target downstream task, \ie, abstractive summarization~\cite{lewis-etal-2020-bart, pmlr-v119-zhang20ae, song2019mass}. Zhang \etal~\cite{zhang-etal-2019-hibert} adopt masked sentence prediction in pre-training stage and employed it as a sentence encoder for extractive summarization on large-scale news corpus. The most similar approach to ours is Wang \etal~\cite{wang-etal-2019-self}, where they introduced three self-supervised pre-training tasks for extractive summarization with CNN/DM datasets~\cite{see-etal-2017-get}. 
Compared to previous work, our approach focuses more on how to incorporate the heterogeneous attributes of a dialogue to self-supervised methods in order to overcome the challenge that fine-tuning is often unstable on small datasets and causes performance degradation~\cite{rothe-etal-2020-leveraging}.

\subsection{Leveraging pre-trained model for text summarization}
Liu \& Lapata~\cite{liu-lapata-2019-text} has shown that BERT can beneficially be applied to both extractive and abstractive document summarization. For abstractive task, their model consists of a BERT pre-trained on extractive summarization task as the encoder and randomly initialized the 6-layer Transformer blocks for the decoder. On the contrary, we aim to leverage the full power of the proposed self-supervised methods by employing BERT as both an encoder and a decoder, which is pre-trained on the task destined to enhance semantic understanding of a dialogue. In recent empirical studies by Google Research~\cite{rothe-etal-2020-leveraging, goodman2019multistage}, it is possible to achieve state-of-the art results on text summarization without any auxiliary task with the encoder-decoder network utilizing pre-trained BERT, RoBERTa, GPT-2 (so-called \bertbert, \bertgpt).
Our implementation of \bertbert~architecture, however, could not have reached higher performance than baselines because the dataset of our choice is much smaller (about 10 times smaller) than previous work. This means that \bertbert~architecture has a great potential for a nice warm-starting model but may not be sufficient for low-resource datasets to achieve a better performance.

\subsection{Abstractive dialogue summarization}
We have chosen a dialogue dataset of messenger-like natural conversations that have very distinct features from formally-written styled documents like a news article. Most salient pieces of conversations are scattered across the utterances by multiple interlocutors, and it makes difficult to decide what the key point of a dialogue is. Moreover, There are no large enough annotated datasets for abstractive dialogue summarization to train deep neural generative models. To address these problems, Ganesh and Dingliwal~\cite{ganesh2020restructuring} propose a two-phase pipeline method that uses discourse labels and an existing document summarizer in the zero-shot learning perspectives.
Feng~\etal~\cite{feng2020incorporating} propose the first to incorporate commonsense knowledge into abstractive dialogue summarization with a graph neural net that includes both utterance and knowledge nodes. Another approach using the graph structure for dialogue summarization is in Zhao~\etal~\cite{zhao-etal-2020-improving}. They have tackled the problem of previous sequence-to-sequence models~\cite{Bahdanau2015NeuralMT} about not paying an attention to handle the sentence-level long-distance dependency and capture the cross-sentence relations by proposing a method that can construct the whole dialogue as a graph for abastractive dialogue summarization. These two approaches constitute the most recent work on the SAMSum corpus. In this work, we choose them as the baselines to compare our model's score with.

\section{Methods}

In this section, we describe how to enhance BERT's semantic understanding of dialogue in self-supervised fashion.

\begin{itemize}
\item \textbf{Switching Utterance}: Similar with previous works~\cite{wu-etal-2019-self, Logeswaran_Lee_Radev_2018} of predicting switched sentences, this task is to predict whether each utterance (not sentence) is switched or not. We switch some utterances selected with the probability \emph{P$_u$}. Hence, the number of switched utterances changes dynamically at each training step.  
At the same time, we mask some interlocutor's name with the probabilty \emph{P$_n$}. This optional constraint of input utterances makes the task more challenging.
\item \textbf{Switching Interlocutor}: For this task, we switch interlocutor's name in utterances instead of utterances with the probabilty \emph{P$_i$} and we did not mask the name of interlocutors. Additionally, we concatenate corresponding reference summary with the utterances for each input sequence, so it can help the model find a mismatch between interlocutor and what they said.
\item \textbf{Inserting Utterance}: For this task, we insert \emph{K} utterances from other dialogues selected with a pre-defined probability into randomly selected \emph{K} positions from inter-utterances and the model predict whether each utterance is from other dialogue or not. If interlocutor's name in inserted utterances remains same as the original, which look obviously unfamiliar, the task has a little worth to solve. Hence, we replace the name of them properly to camouflage where they come from.
\item \textbf{Masking Interlocutor}: This task is similar to masked language modeling~\cite{devlin-etal-2019-bert} task which a model needs to recover masked token from a vocabulary. In our work, we only masked interlocutor's name appeared in reference summary. Then, the model predicts masked names using the information of utterances.
\end{itemize}
\section{Experiments}
\subsection{Dataset}
We evaluate our approach on the SAMSum corpus~\cite{gliwa-etal-2019-samsum} that is constructed by linguists fluent in English, which contains over 16k chat dialogues consists of over 182K utterances (see Table~\ref{tab:samsum} for statistics on dataset).
Each utterance has the specified format that a colon at the beginning of an utterance separates the interlocutor's name and their speaking content.
And also, we append 311 facial emojis into our vocab that appear most often in the train dataset in order to avoid including too many \emph{[unk]} in the input sequence. The effect of appended tokens is on Table~\ref{tab:result}.

\begin{table}[t!]
  \caption{The number of each components of SAMSum corpus, where "Dial." stands for "Dialogue.", "Utter." for "Utterance", "Inter." for "Interlocutor's name" appearing in dialogues. "OOV." is short for the number (proportion) of utterance that contains at least one out of vocabulary token which is mostly facial emojis such as in "Nadine: caaaaaat \emojicat\emojicat\emojicat\emojicat\emojicat\emojicat\emojicat" , and "OOV. w/ FE" for the case "with facial emoji appended vocabulary".}
  \label{tab:samsum}
  \centering
	\resizebox{1.00\columnwidth}{!}{
	\begin{tabular}{l|ccccc}
		\toprule 
		\textbf{Type} & \textbf{\# Dial.} & \textbf{\# Utter.} & \textbf{\# Inter.} & \textbf{\# OOV. } & \textbf{\# OOV. w/ FE} \\
		\midrule
		Train & 14,732 & 164,505 & 4,289 & 3,315 (2.0\%) & 1,165 (0.7\%)\\
		Valid & 819 & 8,860 & 894 & 179 (2.0\%)& 44 (0.4\%)\\
		Test & 818 & 9,212 & 912 & 179 (1.9\%)& 82 (0.9\%)\\
		\bottomrule
	\end{tabular}
	}
\end{table}

\subsection{Implementation and training details}
All our experiment start from publicly available `bert-base-uncased'~version.
For self-supervised tasks, we preprocess the dialogue datasets for our experiments following steps: a) add \emph{[SEP]} token at every end of utterance which improves structural information of dialogue~\cite{zhao-etal-2020-improving} and is used as the utterance representation in our self-supervised methods, b) replace each name with a single token included in our vocabulary to simplify the problem so that model can learn the semantic relations between the subject and action ("Who did What"), rather than distribution of name tokens.
Then, we add a linear layer with dropout (0.1) on top of BERT, where \emph{[SEP]} and \emph{[MASK]} tokens are fed into for classification task.
Throughout abstractive summarization task, we use the shared encoder-decoder architecture (so called \bertshare~ in \cite{rothe-etal-2020-leveraging}) where both model's weights are shared and initialized with BERT trained by one of proposed self-supervised methods. Note that weight sharing between encoder and decoder is necessary rather than optional in that it reduces model parameters and improves the final performance as well in our experimental setup. 

The hyper-parameters of self-supervised learning and summarization are almost identical except training epochs.
We use the AdamW optimizer~\cite{adamw} for both self-supervised and summarization task, with batch size 128, input sequence length 512 and learning rate from 2e-5 to 5e-5, warmup steps are 500.
For self-supervised learning, we train the model until the train loss converged (upper bounded by 5K steps).
After that, we fine-tune the model for summarization task until the validation loss converges.

\begin{table}[t!]
  \caption{Results in terms of ROUGE metric on the SAMSum corpus test set.}
  \label{tab:result}
  \centering
	\resizebox{1.00\columnwidth}{!}{
	\begin{tabular}{lcccc}
		\toprule 
		\textbf{Model} & \textbf{R-1} & \textbf{R-2} & \textbf{R-L} & \textbf{R-AVG} \\ 
		\midrule
		LONGEST-3 & 32.46 & 10.27 & 29.92 & 24.22 \\ 
		Transformer~\cite{46201} & 37.27 & 10.76 & 32.73 & 26.92 \\
		Fast Abs RL Enhanced~\cite{chen-bansal-2018-fast} & 41.95 & 18.06 & 39.23 & 33.08\\
		D-HGN~\cite{feng2020incorporating} & 42.03 & 18.07 & 39.56 & 33.22 \\
		TGDGA~\cite{zhao-etal-2020-improving} & 43.11 & \bf 19.15 & 40.49 & 34.25 \\
		\midrule
		\bertshare~ & 39.07 & 12.74 & 36.05 & 29.29 \\
		+ Facial Emoji & 39.97 & 13.7 & 36.42 & 30.03 \\
		w/ Masking Interlocutor & 40.29 & 13.91 & 37.46 & 30.55 \\
		w/ Inserting Utterance & 44.17 & 18.76 & 41.68 & 34.87 \\
		w/ Switching Interlocutor & 44.01 & 18.03 & 41.42 & 34.49 \\
		w/ Switching Utterance & \bf 44.78 & 19.12 & \bf 42.21 & \bf 35.37 \\
		\bottomrule
	\end{tabular}
	}
\end{table}

\subsection{Evaluation Metrics}
ROUGE~\cite{lin-2004-rouge} is one of standard measures to evaluate machine generated text over many natural language processing fields.
However, the metric based on only n-gram overlapping may not be the best choice for abstractive dialogue summarization~\cite{gliwa-etal-2019-samsum}. Such measure is lacking aspects of fluency, intelligibility, and repetition~\cite{Savelieva2020AbstractiveSO}.
In order to address this issue, we also report consine-similarity between model's prediction (i.e. generated summary) and ground truth using the sentence encoder~\cite{reimers-2019-sentence-bert}, `stsb-roberta-large (available at https://www.sbert.net/)' fine-tuned on Semantic Text Similarity dataset. 
More precisely, we use only the longest 100 samples for calculating cosine-similarity in order to uncover the differences between good and poor quality summaries more clearly.

\begin{table*}[t!]
  \centering
  \caption{An ablation study for two options in three self-supervised methods on SAMSum corpus test set.}
  \label{tab:ablation}
	\begin{tabularx}{0.85\textwidth}{c|c|c|ccccc}
		\toprule 
		\textbf{Method} & \parbox{2cm}{\centering\textbf{Reference}} & \parbox{2cm}{\centering\textbf{Interlocutor}} & \textbf{R-1} & \textbf{R-2} & \textbf{R-L} & \textbf{R-AVG} & \textbf{COS.} \\
		\midrule
		\multirow{2}{*}{Switching Interlocutor}
		& - & \ding{51} & 44.01 & 18.03 & 41.42 & 34.49 &  0.6289 \\
		& \ding{51} & \ding{51} & 42.96 & 17.74& 40.84 & 33.85 &  0.6453 \\
		\midrule
		\multirow{4}{*}{Switching Utterance}
		& \ding{51} & \ding{51} & 44.04 & 18.61 & 41.52 & 34.72 & \textbf{0.6521} \\
		& - & \ding{51} & 44.21 & 18.64 & 41.94 & 34.93 & 0.6396 \\
		& \ding{51} & - & 43.5 & 18.14 & 41.22 & 34.29 & 0.6498 \\
		& - & - & \bf 44.78 & \bf 19.12 & \bf 42.21 & \bf 35.37 & 0.6367 \\
		\midrule
		\multirow{4}{*}{Inserting Utterance}
		& \ding{51} & \ding{51} & 43.38 & 18.04 & 41.02 & 34.15 & 0.6401 \\
		& - & \ding{51} & 44.17 & 18.76 & 41.68 & 34.87 & 0.6292 \\
		& \ding{51} & - & 43.14 & 17.28 & 40.5 & 33.64 & 0.6428 \\
		& - & - & 43.43 & 18.25 & 41.19 & 34.29 & 0.6431 \\
		\bottomrule
	\end{tabularx}
\end{table*}

\section{Results}

\subsection{Evaluation of Self-supervised Learning}

Table \ref{tab:result} shows the result of the proposed self-supervised learning methods on SAMSum dataset. 
The upper part of Table \ref{tab:result} is the scores reported in each paper or \cite{gliwa-etal-2019-samsum}.
We use the official PyRouge package (https://pypi.org/project/py-rouge) to compute the ROUGE score.
In contrast to \bertshare~'s promising results~\cite{rothe-etal-2020-leveraging} on a large corpus of news like CNN/DM, it shows the relatively lower performance on our dataset compared to recent strong baselines. However, all combinations of \bertshare~and our methods improve performance dramatically by up to 6.08\% (Switching Utterance) in averaged ROUGE score.

On the other hand, Masking Interlocutor methods shows the poor result compared to other methods. 
This is because the generative training objective that predicts masked tokens from thousands of candidates usually requires much longer training steps with a larger corpus than other binary decision task (i.e. switched or not), and we will investigate proper training setup in future work.

In addition, we confirm that the inclusion of facial emojis in the vocabulary is advantageous for abstractive summarization task, even though they do not appear in the reference summary at all. This is also evidence that if the encoder is enhanced and is employed as decoder too, it leads to better performance on sequence generation task.

\subsection{Ablation Test}

\subsubsection{Adding reference summary or elimination of interlocutor's name}
In the three among our proposed methods, the reference summary (gold label) and interlocutor's name can be added or removed freely. 
For example, a model can detect the inconsistent utterance order (Switch Utterance) or mismatch of $\langle$interlocutor, utterance$\rangle$ pair (Switch Interlocutor) by only using the information flows in utterances without summary and interlocutor's name.
We conduct ablation studies to investigate how the model incorporates these additional information and the result is on Table \ref{tab:ablation}.
From the result, adding these two factors does not lead to better results in most cases in terms of ROUGE score. 
Note that the highest ROUGE score achieved without any factors.

Interestingly, in most cases, using the concatenation of summary and dialogue as the input sequence for our three methods helps achieve higher scores in terms of cosine-similarity of reference summaries and generated ones.
It implies that there exists inconsistency between ROUGE and cosine-similarity metric and reference summary can help a model learn useful semantic information to generate better summary.

\subsubsection{Probability of Switching and Masking Name}

\begin{figure}[t] 
\begin{center}
\includegraphics[width=1.0\linewidth]{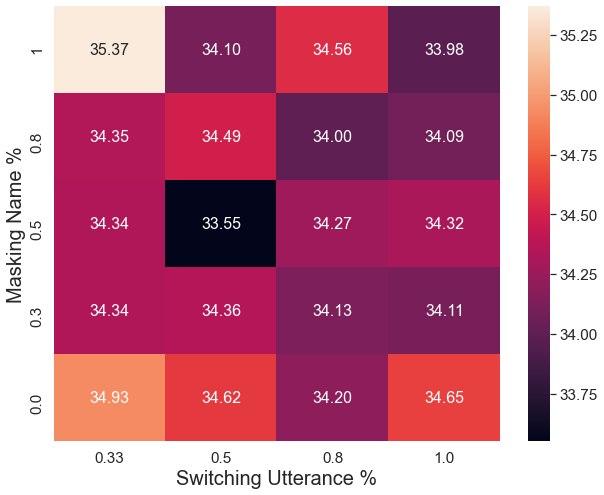}
\end{center}
\caption{Ablation results of Switching Utterance method according to combinations of two probabilities in terms of average ROUGE scores.}
\label{fig:heatmap}
\end{figure}

We also investigate the effects of probability of switching, \emph{P$_u$} and masking name, \emph{P$_n$} for Switching Utterance task.
We set the \emph{P$_u$} a range of [0.33, 1] and (0, 1] for \emph{P$_n$}. 
Even if \emph{P$_u$} is 1.0, that does not mean all utterances are switched, because there are chances to restore the original order of utterances while re-ordering utterances randomly.
The numbers in Figure \ref{fig:heatmap} refer to the average score of ROUGE-1, 2 and L measure, obtained from the model pre-trained by Switching Utterance method with each probability and fine-tuned on SAMSum dataset for abstractive summarization.
As shown in Figure \ref{fig:heatmap}, we found that combination of two probabilities on the edges in the matrix, i.e. (1.0, 1.0), (1.0, 0.0), (0.33, 0), tends to show better performance. 
On the other hand, the combination of 0.5 and 0.5 yields the worst result.

\section{Conclusion}
We proposed four self-supervised methods to enhance semantic understanding of conversational text by multiple interlocutors for abstractive dialogue summarization. The methods, switching utterance, switching interlocutor, inserting utterance, and masking interlocutor, specifically strengthen the learning of structural and component information in dialogues. By enhancing an off-the-shelf pre-trained BERT with our methods, we build an abstractive summarizer in a shared encoder-decoder architecture for sequence-to-sequence training. Our experimental results on the SAMSum corpus indicate a substantive improvement measured in ROUGE scores. Through a careful ablation study, we provide a practical insight for the crucial hyperparameter settings of the proposed self-supervised methods.

 We believe there is still much room for improvements in our setting. We used only `bert-base-uncased'~version to prove our concept, "Can intuitively enhanced representation of encoder produce better performance in the sequence generation task?". Nevertheless, this base model has already achieved promising results so that it will inspire more investigation into engaging large language models with more sophisticated experimental setup.


\bibliographystyle{IEEEtran}
\bibliography{mybib}

\end{document}